\documentclass[10pt,letterpaper,twocolumn]{article}
\usepackage[latin9]{inputenc}
\usepackage{booktabs}
\usepackage{amsmath}
\usepackage{amssymb}
\usepackage{graphicx}
\usepackage[unicode=true,
 bookmarks=false,
 breaklinks=true,pdfborder={0 0 1},backref=section,colorlinks=false]
 {hyperref}

\makeatletter

\pdfpageheight\paperheight
\pdfpagewidth\paperwidth

\providecommand{\tabularnewline}{\\}

\usepackage{cvpr}
\usepackage{times}
\usepackage{epsfig}
\usepackage{graphicx}

\cvprfinalcopy 

\def\cvprPaperID{5576} 

\makeatother

\begin{document}
\title{Learning Regularity in Skeleton Trajectories for Anomaly Detection
in Videos}
\author{Romero Morais$^{1*}$, Vuong Le$^{1}$, Truyen Tran$^{1}$, Budhaditya
Saha$^{1}$, Moussa Mansour$^{2,3}$, Svetha Venkatesh$^{1}$\\
 $^{1}$Applied Artificial Intelligence Institute, Deakin University,
Australia\\
 $^{2}$iCetana, Inc. \textbar{} $^{3}$University of Western Australia,
Australia\\
 \texttt{\small{}$^{1}$\{ralmeidabaratad,vuong.le,truyen.tran,budhaditya.saha,svetha.venkatesh\}@deakin.edu.au}{\small{}}\\
\texttt{\small{}$^{2}$moussa@icetana.com.au}{\small{}
}}
\maketitle
\begin{abstract}
Appearance features have been widely used in video anomaly detection
even though they contain complex entangled factors. We propose a new
method to model the normal patterns of human movements in surveillance
video for anomaly detection using dynamic skeleton features. We decompose
the skeletal movements into two sub-components: global body movement
and local body posture. We model the dynamics and interaction of the
coupled features in our novel Message-Passing Encoder-Decoder Recurrent
Network. We observed that the decoupled features collaboratively interact
in our spatio-temporal model to accurately identify human-related
irregular events from surveillance video sequences. Compared to traditional
appearance-based models, our method achieves superior outlier detection
performance. Our model also offers ``open-box'' examination and
decision explanation made possible by the semantically understandable
features and a network architecture supporting interpretability.
\end{abstract}

\section{Introduction}

\begin{figure}
\begin{centering}
\includegraphics[width=1\columnwidth]{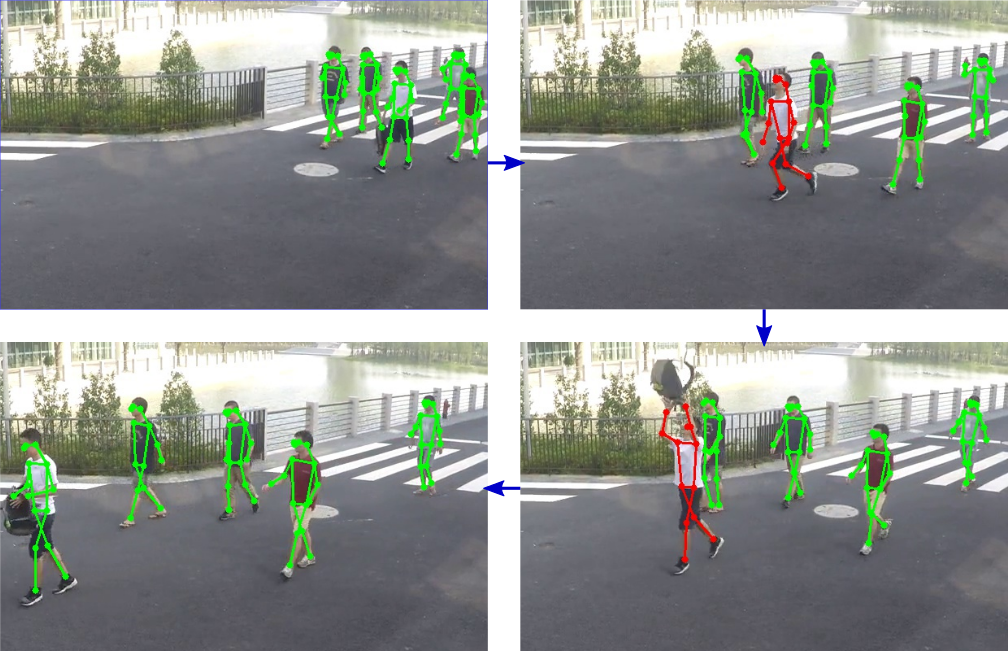}
\par\end{centering}
\caption{We detect human-related anomalies in video by learning regular spatio-temporal
patterns of skeleton features. In this example, we detect the anomalous
event of a person catching a backpack. This anomaly is detected by
using his unusual skeleton pose and motion compared to those of normal
activities. Skeletons in red denote high anomaly scores, while skeletons
in green denote low anomaly scores. The order of the frames is specified
by the blue arrows.\label{fig:intro}}

\end{figure}
Video anomaly detection is a core problem of unsupervised video modeling.
An effective solution is learning the regular patterns in normal training
video sequences in an unsupervised setting, based on which irregular
events in test videos can be detected as outliers. The problem is
challenging due to the lack of human supervision and the ambiguous
definition of human-perceivable abnormality in video events. Most
current approaches operate on pixel-based appearance and motion features.
These features are usually extracted from whole frames \cite{chong2017abnormal,hasan2016learning,liu2018anopred,luo2017revisit,xu2017detecting},
localized on a grid of image patches \cite{vu2017energy}, or concentrated
on pre-identified regions \cite{cocsar2017toward,hinami2017joint}.
Unfortunately, pixel-based features are high-dimensional unstructured
signals sensitive to noise, that mask important information about
the scene \cite{zimek2012survey}. Furthermore, the redundant information
present in these features increases the burden on the models trained
on them to discriminate between signal and noise. 

Another key limitation of current methods is the lack of interpretability
due to the semantic gap between visual features and the real meaning
of the events. This limitation can be amplified through processing
in deep neural networks \cite{bau2017network}. This lack of understanding
prevents practitioners from using domain knowledge to customize model
architectures and obstructs error analysis.

In this paper, we propose to leverage 2D human skeleton trajectories
for detecting abnormal events related to human behavior in surveillance
videos. The skeleton trajectories contain the locations of a collection
of body joints in the spatio-temporal domain of video sequences, as
illustrated in Figure \ref{fig:intro}. By using skeleton features
we explicitly exploit the common structure of surveillance videos,
which consists of humans and objects attached to them moving on top
of a static background. Compared to appearance-based representations,
skeleton features are compact, strongly structured, semantically rich,
and highly descriptive about human action and movement, which are
keys to anomaly detection. 

By studying the human skeleton dynamics in a large collection of surveillance
videos, we observed that human behavioral irregularity can be factorized
into a few factors regarding body motion and posture, such as location,
velocity, direction, pose and action. Motivated by this natural factorization,
we propose to decompose the dynamic skeleton motions into two sub-processes,
one describing global body movement and the other local body posture.
The global movement tracks the dynamics of the whole body in the scene,
while the local posture describes the skeleton configuration in the
canonical coordinate frame of the body's bounding box, where the global
movement has been factored out.

We jointly model the two sub-processes in a novel model called Message-Passing
Encoder-Decoder Recurrent Neural Network (MPED-RNN). The network consists
of two RNN branches dedicated to the global and local feature components.
The branches process their data separately and interact via cross-branch
message-passing at each time step. The model is trained end-to-end
and regularized so that it distills the most compact profile of the
normal patterns of training data and effectively detects abnormal
events. In addition to anomaly detection, MPED-RNN supports open-box
interpretation of its internal reasoning by providing the weights
of the contributing factors to the decision and the visualization
of these factors. We trial our method on two of the most challenging
video anomaly datasets and compare our results with the state-of-the-art
on the field. The results show that our proposed method is competitive
in detection performance and easier to analyze the failure modes.

\section{Related Work}

\subsection{Video anomaly detection}

Unsupervised video anomaly detection methods have been an old-timer
topic in the video processing and computer vision communities. Traditional
approaches consider video frames as separate data samples and model
them using one-class classification methods, such as one-class SVM
\cite{xu2017detecting} and mixture of probabilistic PCA \cite{kim2009observe}.
These methods usually attain suboptimal performance when processing
large scale data with a wide variety of anomaly types. 

Recent approaches rejuvenate the field by using convolutional neural
networks (CNN) to extract high-level features from video frame intensity
and achieve improved results. Some of these methods include Convolutional
autoencoder \cite{hasan2016learning}, spatio-temporal autoencoder
\cite{chong2017abnormal}, 3D Convnet AE \cite{zhao2017spatio}, and
Temporally-coherent Sparse Coding Stacked-RNN \cite{luo2017revisit}.
Acknowledging the limitation of the intensity based features such
as sensitivity to appearance noise, Liu \etal \cite{liu2018anopred}
proposed to use the prediction of optical flow in their temporal coherent
loss, effectively filtering out parts of the noise in pixel appearance.
However, optical flow is costly to extract and still far from the
semantic nature of the events.

Structured representations have recently attracted increased attention
for its potential to get closer to the semantic concepts present in
the anomalies. In \cite{yuan2015online}, object trajectories were
used to guide the pooling of the visual features so that interesting
areas are paid more attention to. Towards model interpretability,
Hinami \etal \cite{hinami2017joint} proposed to use object, attribute,
and action detection labels to understand the reason of abnormality
scores. Although it works well for a number of events, their method
fails in many cases due to the incompleteness of the label sets and
the distraction from unrelated information in the labels.

Our method of using skeleton features is another step towards using
low-dimensional semantic-rich features for anomaly detection. We also
advance the research efforts toward interpretability of the anomaly
detection models by providing the ability to explain every abnormal
event in our factorized semantic space.

\subsection{Human trajectory modeling}

Human motion in video scenes is an important factor for studying social
behavior. It has been applied in multiple computer vision applications,
mostly with supervised learning tasks such as action recognition \cite{du2016representation}
and person re-identification \cite{elaoud2017analysis}. Recently,
more effort has been invested into unsupervised learning of human
motion in social settings \cite{alahi2016social,gupta2018social,villegas2017learning}
and single pose configuration \cite{fragkiadaki2015recurrent}. In
this work, we propose to expand the application of skeleton motion
features to the task of video anomaly detection. In MPED-RNN, we share
the encoder-decoder structure with most unsupervised prediction models.
However, instead of perfectly generating the expected features, we
aim at distilling only the principal feature patterns so that anomalies
are left out. This involves building a highly regulated autoencoder.

Regarding feature representation and modeling, apart from traditional
methods that rely on hand-crafted local features and state machines,
several recent works proposed to use interacting recurrent networks
for motion modeling in social settings \cite{alahi2016social,gupta2018social}.
In these approaches, the input data are in the form of whole body
xy-location sequences while local postures are ignored. To bridge
this gap, in \cite{fragkiadaki2015recurrent,villegas2017learning}
the input of the RNN is extended to the skeleton joint locations.
Recently, Du \etal \cite{du2016representation} proposed to divide
the skeleton joints into five parts, which are jointly modeled in
a five-branch bidirectional neural network. Different to previous
approaches, we factorize skeleton motion based on natural decomposition
of human motion into global movement/local deformation and model them
jointly in an interactive recurrent network.

\section{Method}

The anomalous human-related events in a surveillance video scene can
be identified by the irregular human movement patterns observed in
the video. Our method detects those anomalies by learning a regularity
model of the dynamic skeleton features found in training videos. We
assume the skeleton trajectories have already been extracted from
the videos. At each time step $t$, a skeleton is represented by a
set of joint locations in image coordinates $f_{t}=\left(x_{t}^{i},y_{t}^{i}\right)_{i=1..k}$,
where $k$ is the number of skeleton joints. This set of temporal
sequences is the input to our anomaly detection algorithm.

\subsection{Skeleton Motion Decomposition\label{subsec:Skeleton-Motion-Decomposition}}

\begin{figure}[h]
\begin{centering}
\includegraphics[width=0.95\columnwidth]{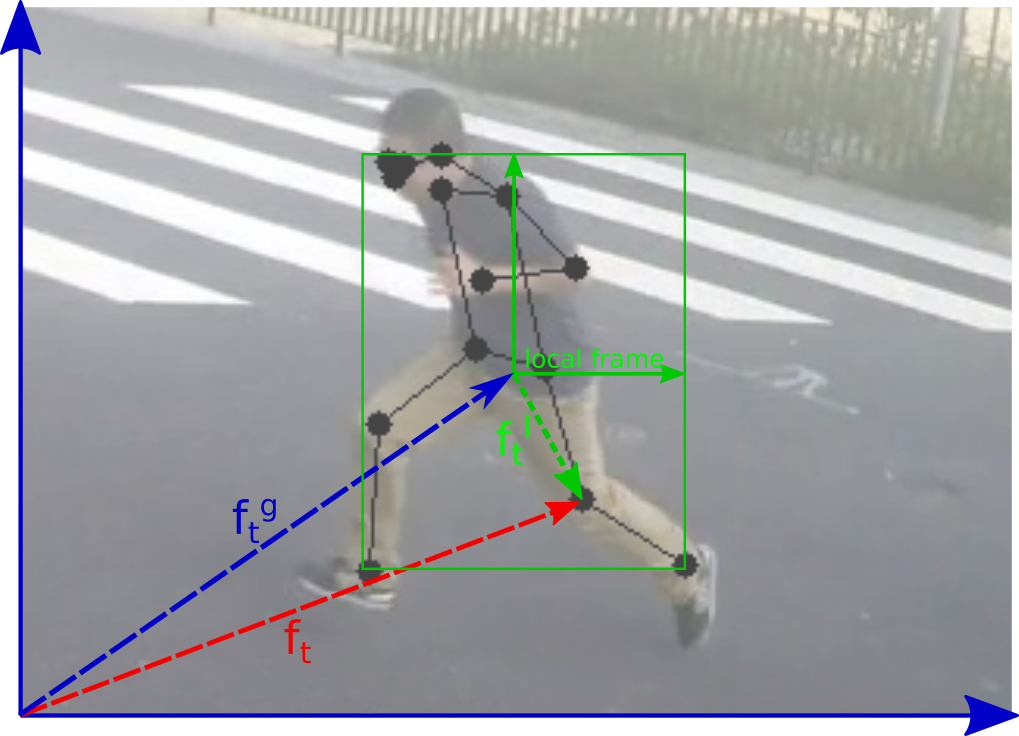}
\par\end{centering}
\caption{Global and local decomposition of a skeleton in a frame. Based on
the canonical local reference frame defined by the green bounding
box, the location vector of the left knee joint$f_{t}$ (dashed red)
is decomposed into global $f_{t}^{g}$(dashed blue) and local $f_{t}^{l}$
(dashed green) components. The bounding box's width and height are
included as global features and used to normalize local features.}

\label{fig:global_local_decomposition}
\end{figure}
Naturally, human motion consists of two factors: rigid movement of
the whole body and non-rigid deformation of the skeleton joints. The
simplest way to model human motion using a recurrent network is by
feeding it the raw sequence of skeleton trajectories in image coordinates
$f_{t}=\left(x_{t}^{i},y_{t}^{i}\right)$ \cite{fragkiadaki2015recurrent,villegas2017learning},
implicitly merging the global and local factors together. This solution
performs well in videos with uniform skeleton scales and types of
activities where the contribution of the two factors is consistent.
On realistic surveillance videos, however, the scales of human skeletons
vary largely depending on their location and actions. For skeletons
in the near field, the observed motion is mainly influenced by the
local factor. Meanwhile, for skeletons in the far field, the motion
is dominated by the global movement while local deformation is mostly
ignored. 

Inspired by the natural composition of human skeleton motion and motivated
by the factorization models widely used in statistical modeling, we
propose to decompose the skeletal motion into ``global'' and ``local''
components. The global component carries information about the shape,
size and rigid movement of the human bounding box. The local component
models the internal deformation of the skeleton and ignores the skeleton's
absolute position in relation to the environment. 

Geometrically, we set a canonical reference frame attached to the
human body (called \emph{local frame}), which is rooted at the center
of the skeleton's bounding box. The global component is defined as
the absolute location of the local frame center within the original
image frame. On the other hand, the local component is defined as
the residue after subtracting the global component from the original
motion. It represents the relative position of skeleton joints with
respect to the bounding box. This decomposition is illustrated in
Figure~\ref{fig:global_local_decomposition} and can be written in
2D vector space as: 
\begin{equation}
f_{t}^{i}=f_{t}^{g}+f_{t}^{l,i}\label{eq:decompose_vector}
\end{equation}

In 2D image space, xy-coordinates alone poorly represent the real
location in the scene because the depth is missing. However, the size
of a skeleton's bounding box is correlated with the skeleton's depth
in the scene. To bridge this gap, we augment the global component
with the width and height of the skeleton's bounding box $f^{g}=(x^{g},y^{g},w,h)$
and use them to normalize the local component $f^{l,i}=(x^{l,i},y^{l,i})$.
These features can be calculated from the input features as:
\begin{align}
x^{g}=\frac{max(x^{i})+min(x^{i})}{2}; & \quad y^{g}=\frac{max(y^{i})+min(y^{i})}{2}\nonumber \\
w=max(x^{i})-min(x^{i}); & \quad h=max(y^{i})-min(y^{i})\label{eq:decompose_global}\\
x^{l,i}=\frac{x^{i}-x^{g}}{w}; & \quad y^{l,i}=\frac{y^{i}-y^{g}}{h}\label{eq:decompose_local}
\end{align}

The global and local dynamics can be modeled separately as two concurrent
sub-processes. In generic videos, these two processes can even manifest
independently. For example, a person can move her limbs around while
keeping her global location relatively still. Similarly, a person
riding a motorbike can move around while having a relatively fixed
pose. However, given a specific context, regular human activities
contain a strong correlation between these two components. Therefore
breaking the cross-component correlation is also a sign of abnormality.
In the previous examples, if those actions occurred in the scene where
people were normally walking, they would be valid anomaly events.
In the next section, we present how both individual dynamic patterns
and the relationship between these two components are modeled in our
MPED-RNN model.
\begin{center}
\begin{figure*}[h]
\centering{}\includegraphics[width=0.95\textwidth]{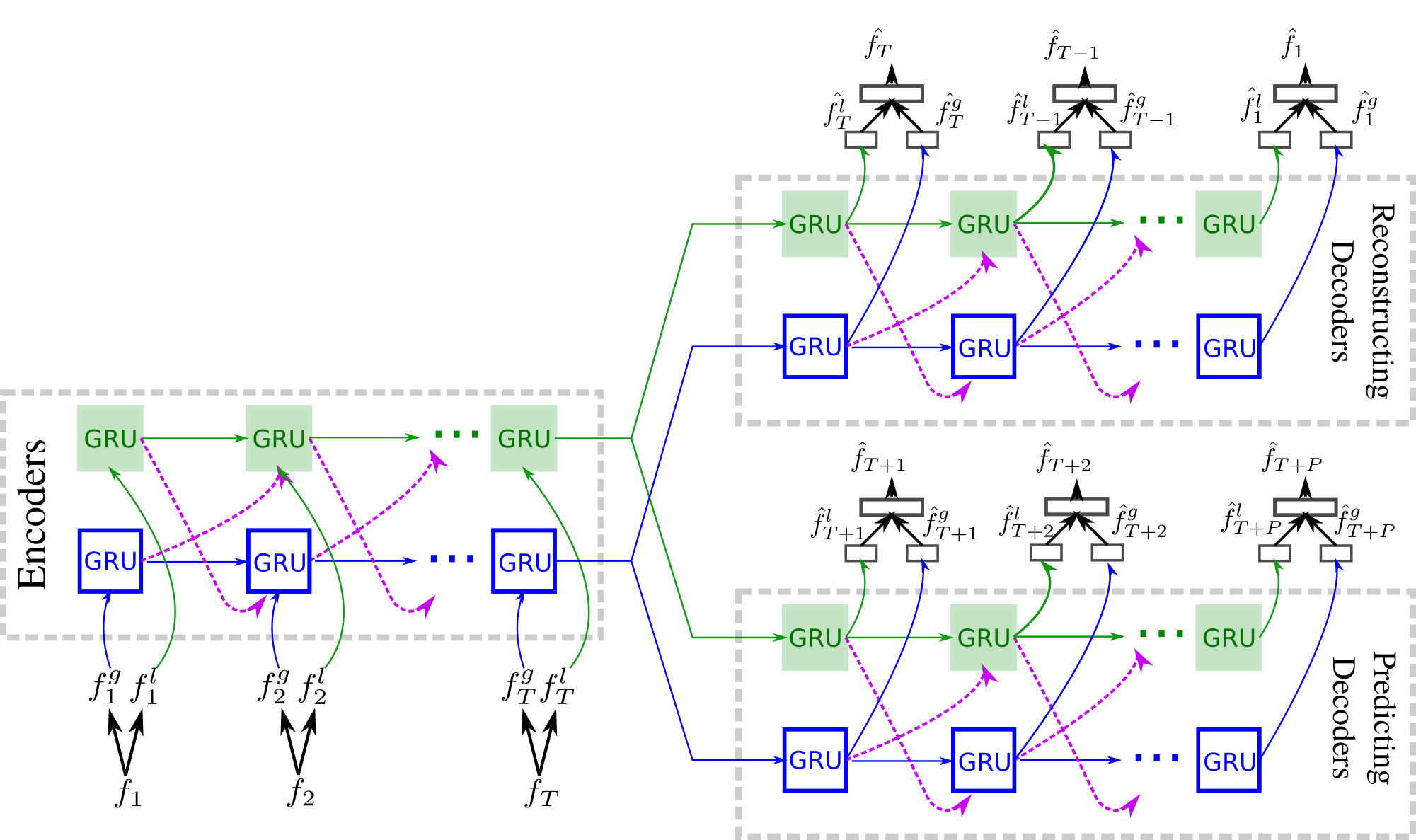}\caption{MPED-RNN consists of two interacting branches for two skeleton feature
components. The local branch is drawn in green with shaded GRU blocks
and the global branch is drawn in blue with transparent GRU blocks.
The two components interact through messages (purple dashed) exchanged
between the branches. The outputs are generated by a set of MLPs,
represented by black rectangles.\label{fig:comprae_architecture}}
\end{figure*}
\par\end{center}

\subsection{MPED-RNN Architecture}

MPED-RNN models the global and local components as two interacting
sub-processes where the internal state of one process is used as extra
features to the input of the other process. More specifically, the
model consists of two recurrent encoder-decoder network branches,
each of them dedicated to one of the components. Each branch of the
model has the single-encoder-dual-decoder architecture with three
RNNs: Encoder, Reconstructing Decoder and Predicting Decoder. This
structure is similar to the composite LSTM autoencoder (LSTM AE) of
Srivastava \etal \cite{srivastava2015unsupervised}. However, unlike
LSTM AE, MPED-RNN does not only model the dynamics of each individual
component, but also the interdependencies between them through a cross-branch
message-passing mechanism. We use Gated Recurrent Units (GRU) \cite{cho2014learning}
in every segment of MPED-RNN for its simplicity and similar performance
to LSTM \cite{greff2017lstm}. At each time step, the GRU unit of
one branch receives a message from the other branch informing its
internal state at the previous time step. This message is incorporated
into the GRU structure by treating it as an additional input. The
same procedure is applied to the other branch. MPED-RNN's architecture
is depicted in Figure~\ref{fig:comprae_architecture}.

Given an input skeleton segment of length $T$, we first initialize
the hidden states of all the GRUs to null. Then, for each time step
$t$, the skeleton $f_{t}$ is decomposed into $f_{t}^{g}$ and $f_{t}^{l}$
(using Eqs.~(\ref{eq:decompose_vector}), (\ref{eq:decompose_global})
and (\ref{eq:decompose_local})), which are input to the global encoder
($E^{g}$) and to the local encoder ($E^{l}$), respectively. The
messages to be exchanged between the global and local branches are
computed as specified by Eqs. \ref{eq:msg_local_to_global} and \ref{eq:msg_global_to_local}
below.
\begin{align}
m_{t}^{l\rightarrow g} & =\sigma\big(W^{l\rightarrow g}h_{t-1}^{l}+b^{l\rightarrow g}\big)\label{eq:msg_local_to_global}\\
m_{t}^{g\rightarrow l} & =\sigma\big(W^{g\rightarrow l}h_{t-1}^{g}+b^{g\rightarrow l}\big)\label{eq:msg_global_to_local}
\end{align}

For $t=1,2,\ldots,T$, the global and local segments are encoded using
Eqs. \ref{eq:global_enc} and \ref{eq:local_enc}: 
\begin{align}
E^{ge}:h_{t}^{ge} & =\textrm{GRU}\left(\left[f_{t}^{g},m_{t}^{le\rightarrow ge}\right],h_{t-1}^{ge}\right)\label{eq:global_enc}\\
E^{le}:h_{t}^{le} & =\textrm{GRU}\left(\left[f_{t}^{l},m_{t}^{ge\rightarrow le}\right],h_{t-1}^{le}\right)\label{eq:local_enc}
\end{align}

After encoding the input segments, the global and local Reconstructing
Decoders initialize their hidden states as $h_{T}^{gr}=h_{T}^{ge}$
and $h_{T}^{lr}=h_{T}^{le}$, respectively, and for $t=T,T-1,...,1$,
we have:
\begin{align}
D_{r}^{g} & :h_{t-1}^{gr}=\textrm{GRU}(m_{t}^{lr\rightarrow gr},h_{t}^{gr})\label{eq:rec_global}\\
D_{r}^{l} & :h_{t-1}^{lr}=\textrm{GRU}(m_{t}^{gr\rightarrow lr},h_{t}^{lr})\label{eq:rec_local}
\end{align}

Similarly, the global and local Predicting Decoders initialize their
hidden states as $h_{T}^{gp}=h_{T}^{ge}$ and $h_{T}^{lp}=h_{T}^{le}$,
respectively, and for $t=T+1,T+2,\ldots,T+P$, we have:
\begin{align}
D_{p}^{g} & :h_{t}^{gp}=\textrm{GRU}(m_{t}^{lp\rightarrow gp},h_{t-1}^{gp})\label{eq:pred_global}\\
D_{p}^{l} & :h_{t}^{lp}=\textrm{GRU}(m_{t}^{gp\rightarrow lp},h_{t-1}^{lp})\label{eq:pred_local}
\end{align}

In training, the dual decoders in the MPED-RNN's architecture jointly
enforce the encoder to learn a compact representation rich enough
to reconstruct its own input and predict the unseen future. Meanwhile,
in testing, the abnormal patterns cannot be properly predicted because
they were neither seen before nor follow the normal dynamics.

In each decoder network, the projected features of the corresponding
decoders, $\hat{f_{t}^{g}}$ and $\hat{f_{t}^{l}}$, are independently
generated from the hidden states $h_{t}^{g}$ and $h_{t}^{l}$ by
fully-connected layers. These two projected features are concatenated
and input to another fully-connected layer, which generates the projected
perceptual feature $\hat{f}_{t}$ in the original image space. Ideally,
$\hat{f_{t}}$ can be calculated from $\hat{f_{t}^{g}}$ and $\hat{f_{t}^{l}}$
by inverting Eqs.~(\ref{eq:decompose_global}) and (\ref{eq:decompose_local}).
However, by being projections into low-dimensional subspaces, a direct
computation is unlikely to be optimal. Thus, using a fully-connected
layer to learn the inverse mapping allows the computation to be robust
to noise. These projected features are used to evaluate the conformity
of an input sequence of skeletons to the learned normal behavior and
hence are used to build the loss function for training and score function
for testing. These procedures are detailed next.

\subsection{Training MPED-RNN\label{subsec:Training-MPED-RNN}}

\paragraph*{Training setup}

The trajectory of a person can span many frames in a video. However,
recurrent networks are trained on fixed-size sequences. To cope with
this issue, we extract fixed-size segments from every skeleton's trajectory
using a sliding-window strategy. Therefore, each segment is computed
as:
\begin{equation}
\textrm{seg}_{i}=\{f_{t}\}_{t=b_{i}..e_{i}}\label{eq:sliding_window}
\end{equation}
where $b_{i}$ and $e_{i}$ are beginning and ending indices of the
$i$-th segment calculated from the chosen sliding stride $s$ and
segment length $T$:
\begin{equation}
b_{i}=s\times i;\:e_{i}=s\times i+T\label{eq:window_ends}
\end{equation}

During training, batches of training segments are decomposed into
global and local features, which are input to MPED-RNN. 

\paragraph*{Loss functions}

We consider three loss functions defined in three related coordinate
frames. The Perceptual loss $L_{p}$ constrains MPED-RNN to produce
the normal sequences in the image coordinate system. The Global loss
$L_{g}$ and the Local loss $L_{p}$ act as regularization terms that
enforce that each encoder-decoder branch of MPED-RNN work as designed.
Each of the losses includes the mean squared error made by the reconstructing
and predicting decoders:
\begin{equation}
L_{*}(\textrm{seg}_{i})=\frac{1}{2}\left(\frac{1}{T}\sum_{t=b_{i}}^{e_{i}}\left\Vert \hat{f_{t}^{*}}-f_{t}^{*}\right\Vert _{2}^{2}+\frac{1}{P}\sum_{t=e_{i}+1}^{e_{i}+P}\left\Vert \hat{f_{t}^{*}}-f_{t}^{*}\right\Vert _{2}^{2}\right)\label{eq:loss_calculation}
\end{equation}
where $P$ denotes the prediction length and $*$ represents one of
$l$, $g$ or $p$. In case of $p$ notice that it makes $f_{t}^{p}$
equal to $f_{t}$ of Section \ref{subsec:Skeleton-Motion-Decomposition}.
The prediction loss is truncated if the end of trajectory is reached
within the prediction length.

The three losses contribute to the combined loss by a weighted sum:
\begin{equation}
L(\textrm{seg}_{i})=\lambda_{g}L_{g}(\textrm{seg}_{i})+\lambda_{l}L_{l}(\textrm{seg}_{i})+\lambda_{p}L_{p}(\textrm{seg}_{i})\label{eq:loss}
\end{equation}
where $\left\{ \lambda_{g},\lambda_{l},\lambda_{p}\right\} \ge0$
are corresponding weights to the losses. 

In training, we minimize the combined loss in Eq.~(\ref{eq:loss})
by optimizing the parameters of GRU cells of the RNN networks, message
building transformations in Eqs. \ref{eq:msg_local_to_global} and
\ref{eq:msg_global_to_local}, and the output MLPs. 

\paragraph*{Model regularization}

When training autoencoder style models for anomaly detection, a major
challenge is that even if the model learns to generate normal data
perfectly, there is still no guarantee that the model will produce
high errors for abnormal sequences \cite{liu2018anopred}. In training
MPED-RNN, we address this challenge by empirically searching for the
smallest latent space that still adequately covers the normal patterns
so that outliers fall outside the manifold represented by this subspace.

We implement this intuition by splitting the normal trajectories into
training and validation subsets, and use them to regularize the network's
hyperparameters that govern the capacity of the model (\eg number
of hidden units). More specifically, we train a high capacity network
and record the lowest loss on the validation set. The validation set
is also used for early stopping. Then, we train a network with lower
capacity and record the lowest loss on the validation set again. We
repeat this procedure until we find the network with the smallest
capacity that is still within 5\% of the initial validation loss attained
by the high capacity network.

\subsection{Detecting Video Anomalies}

To estimate the anomaly score of each frame in a video, we follow
a four-step algorithm:
\begin{enumerate}
\item \emph{Extract segments}: With each trajectory, we select the overlapping
skeleton segments by using a sliding window of size $T$ and stride
$s$ on the trajectory, similar to Eqs.~(\ref{eq:sliding_window})
and (\ref{eq:window_ends}). 
\item \emph{Estimate segment losses}: We decompose the segment using Eq.~(\ref{eq:decompose_vector})
and feed all segment features to the trained MPED-RNN, which outputs
the normality loss as in Eq.~(\ref{eq:loss}).
\item \emph{Gather skeleton anomaly score: }To measure the conformity of
a sequence to the model given both the past and future context, we
propose a voting scheme to gather the losses of related segments into
an anomaly score for each skeleton instance:
\begin{equation}
\alpha_{f_{t}}=\frac{\sum_{u\in S_{t}}L_{p}(u)}{|S_{t}|}\label{eq:instance_loss}
\end{equation}
 where $S_{t}$ denotes the set of decoded segments that contain $f_{t}$
from both reconstruction and prediction. For each of those segments
$u$, the corresponding perceptual loss, $L_{p}(u)$, is calculated
by Eq.~(\ref{eq:loss_calculation}).
\item \emph{Calculate frame anomaly score:} The anomaly score of a video
frame $v_{t}$ is calculated from the score of all skeleton instances
appearing in that frame by a max pooling operator:
\begin{equation}
\alpha_{v_{t}}=max\left(\alpha_{f_{t}}\right)_{f_{t}\in\textrm{Skel}(v_{t})}\label{eq:frame_loss}
\end{equation}
where $\textrm{Skel}(v_{t})$ stands for the set of skeleton instances
appearing in the frame. The choice of max pooling over other aggregation
functions is to suppress the influence of normal trajectories present
in the scene, since the number of normal trajectories can vary largely
in real surveillance videos. We then use $\alpha_{v_{t}}$ as the
frame-level anomaly score of $v_{t}$ and use it to calculate all
accuracy measurements. 
\end{enumerate}

\subsection{Implementation Details}

To detect skeletons in the videos, we utilized Alpha Pose \cite{fang2017rmpe}
to independently detect skeletons in each video frame. To track the
skeletons across a video, we combined sparse optical flow with the
detected skeletons to assign similarity scores between pairs of skeletons
in neighboring frames, and solved the assignment problem using the
Hungarian algorithm \cite{kuhn1955hungarian}. The global and local
components of the skeleton trajectories are standardized by subtracting
the median of each feature, and scaling each feature relative to the
10\%-90\% quantile range. All recurrent encoder-decoder networks have
similar architectures but are trained with independent weights. The
regularization of MPED-RNN's hyperparameters is done for each data
set, following the method described in Section \ref{subsec:Training-MPED-RNN}.

\section{Experiments}

We evaluate our method on two datasets for video anomaly detection:
ShanghaiTech Campus \cite{luo2017revisit} and CUHK Avenue \cite{lu2013abnormal}.
Each of these datasets has specific characteristics in terms of data
source, video quality and types of anomaly. Therefore, we setup customized
experiments for each of them.

\subsection{ShanghaiTech Campus Dataset}

The ShanghaiTech Campus dataset \cite{luo2017revisit} is considered
one of the most comprehensive and realistic datasets for video anomaly
detection currently available. It combines footage of 13 different
cameras around the ShanghaiTech University campus with a wide spectrum
of anomaly types. Because of the sophistication of the anomaly semantics,
current methods struggle to get adequate performance on it. 

Most of the anomaly events in the ShanghaiTech dataset are related
to humans, which are the target of our method. We left out 6/107 test
videos whose abnormal events were not related to humans and kept the
other 101 videos as a subset called Human-related (HR) ShanghaiTech.
Most of the experiments discussed in this section are conducted on
the HR-ShanghaiTech dataset.

\subsubsection{Comparison with Appearance-based Methods\label{subsec:Comparison-with-Appearance-based}}

\begin{table}
\caption{Frame-level ROC AUC performance of MPED-RNN and other state-of-the-art
methods on the ShanghaiTech dataset and its human-related subset.
We use the reported results of the referenced methods on ShanghaiTech
and carry out their identical experiments on HR-ShanghaiTech whenever
possible.\label{tab:Performances-SHT}}

\centering{}%
\begin{tabular}{clcc}
\toprule 
\multicolumn{2}{c}{} & HR-ShanghaiTech & ShanghaiTech\tabularnewline
\midrule
\multicolumn{2}{l}{Conv-AE \cite{hasan2016learning}} & 0.698 & 0.704\tabularnewline
\addlinespace
\multicolumn{2}{l}{TSC sRNN \cite{luo2017revisit}} & N/A & 0.680\tabularnewline
\addlinespace
\multicolumn{2}{l}{Liu \etal \cite{liu2018anopred}} & 0.727 & 0.728\tabularnewline
\addlinespace
\midrule 
\multicolumn{2}{l}{MPED-RNN} & \textbf{0.754} & \textbf{0.734}\tabularnewline
\bottomrule
\addlinespace
\end{tabular}
\end{table}
We train MPED-RNN on all training videos, which is the practice adopted
in previous works. Table \ref{tab:Performances-SHT} compares the
frame-level ROC AUC of MPED-RNN against three state-of-the-art methods.
We observe that on HR-ShanghaiTech, MPED-RNN outperforms all the compared
methods. For completeness, we also evaluate MPED-RNN on the original
dataset where non-human related anomalies are present and MPED-RNN
still attains the highest frame-level ROC AUC.

\begin{figure}

\begin{centering}
\includegraphics[width=0.95\columnwidth]{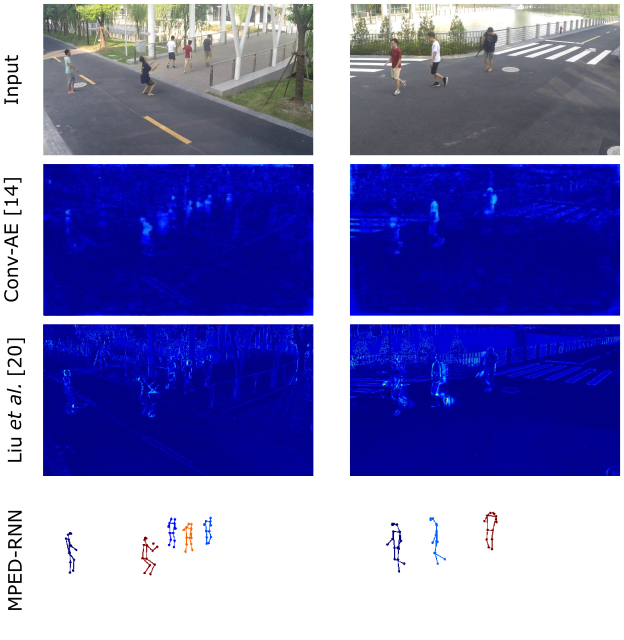}
\par\end{centering}
\caption{\label{fig:comparison_between_methods} Anomaly score map of Conv-AE
\cite{hasan2016learning}, Liu \etal \cite{liu2018anopred} and MPED-RNN
in jet color map. Higher scores are represented closer to red while
lower scores are represented closer to blue. The first row shows the
original input frames and subsequent rows show the score map of each
method. Since MPED-RNN focuses on skeletons it does not produce any
score on background pixels.}

\end{figure}

To understand how the detection of anomalies is made by all models,
we visually compare in Figure \ref{fig:comparison_between_methods}
the map of anomaly scores produced by MPED-RNN to those produced by
Conv-AE \cite{hasan2016learning} and Liu \etal \cite{liu2018anopred}.
As we can observe, our method avoids many irrelevant aspects of the
scene since we focus on skeletons. On the other hand, the other two
methods try to predict the whole scene and are more susceptible to
noise in it.

\begin{center}
\begin{figure*}
\begin{centering}
\includegraphics[width=0.9\textwidth]{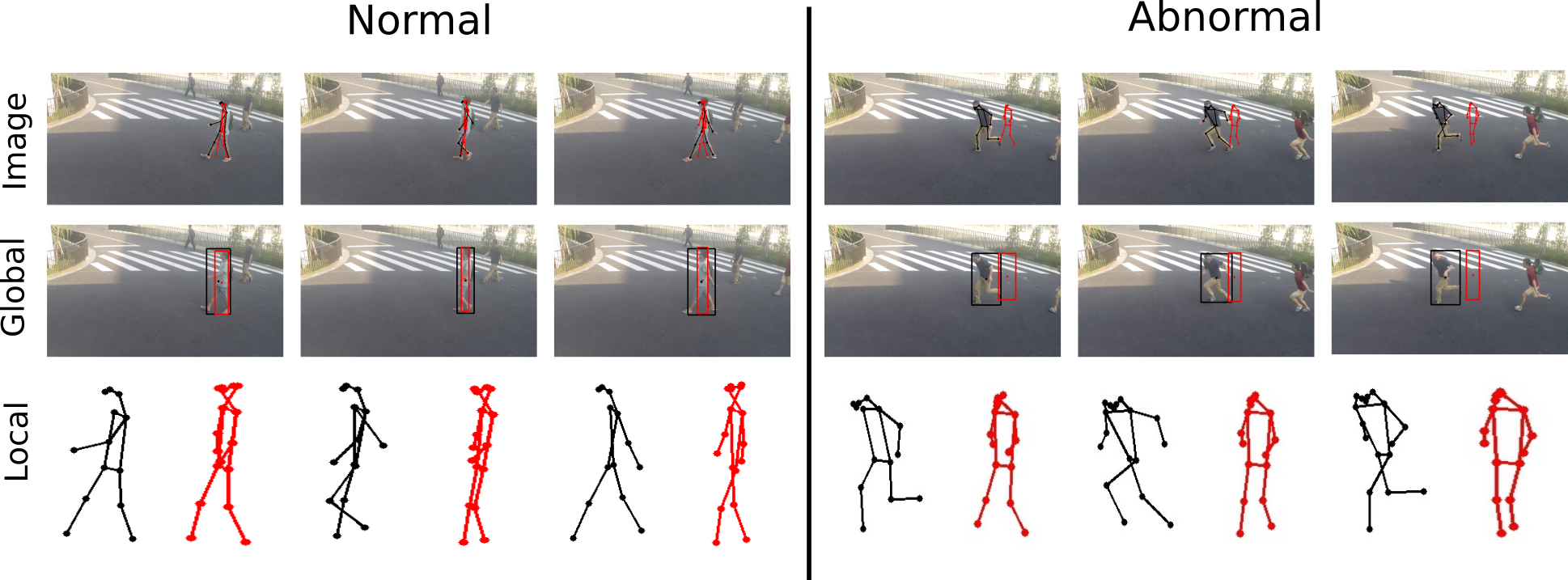}
\par\end{centering}
\caption{\label{fig:Visualization-of-the-generated-data}Visualization of the
predicted features (red) compared to input features (black) in a sample
case for a normal trajectory and an anomalous trajectory.}
\end{figure*}
\par\end{center}

\subsubsection{Interpreting Open-box MPED-RNN}

For a deeper understanding on how MPED-RNN works under the hood, we
visualize the features generated from the global and local predicting
decoders, and the predicted skeleton in image space. For comparison,
we also draw the corresponding features of the input sequence. Figure
\ref{fig:Visualization-of-the-generated-data} shows two example sequences
from the same scene, a normal example and an anomalous example. This
scene is of the walking area in the campus, where regular activities
include people standing and walking casually. In the normal sequence,
the predictions follow the input closely in all three domains, which
shows that MPED-RNN encodes enough information to predict the normal
sequence. On the other hand, the abnormal event contains a person
running. Its predicted global bounding box lags behind the input bounding
box, indicating that the expected movement is slower than the observed
one. The local prediction also struggles to reproduce the running
pose and ends up predicting a walking gait remotely mimicking the
original poses.

\subsubsection{Ablation Study}

Table~\ref{tab:ablation_study} reports the results of simplified
variants of MPED-RNN. It confirms that RNN is needed for this problem,
and when both global and local sub-processes are modeled, message
passing between the sub-processes is necessary. It also shows that
the dual decoders are valuable for regularizing the model and detecting
anomalies.

\begin{table}
\caption{Ablation study about the components of MPED-RNN. We show the frame-level
ROC AUC of simpler models that compose MPED-RNN on the HR-ShanghaiTech
dataset. AE: Frame-level Autoencoder, ED: Encoder-Decoder, G+L: Global
and Local features without message passing. The columns stand for
different ways the loss is calculated; Rec: reconstruction only, Pred:
prediction ony, Rec+Pred: reconstruction and prediction combined.
\label{tab:ablation_study}}

\centering{}%
\begin{tabular}{lccc}
\toprule 
 & \multicolumn{3}{c}{HR-ShanghaiTech}\tabularnewline
\midrule 
 & Rec. & Pred. & Rec. + Pred.\tabularnewline
\midrule 
AE/Image & 0.674 & N/A & N/A\tabularnewline
\addlinespace
ED-RNN/Global & 0.680 & 0.688 & 0.689\tabularnewline
\addlinespace
ED-RNN/Local & 0.700 & 0.714 & 0.715\tabularnewline
\addlinespace
ED-RNN/G+L & 0.699 & 0.722 & 0.713\tabularnewline
\midrule
\addlinespace
MPED-RNN & \textbf{0.744} & \textbf{0.745} & \textbf{0.754}\tabularnewline
\bottomrule
\addlinespace
\end{tabular}
\end{table}

\subsubsection{Error Mode Analysis}

\begin{figure}
\begin{centering}
\includegraphics[width=0.95\columnwidth]{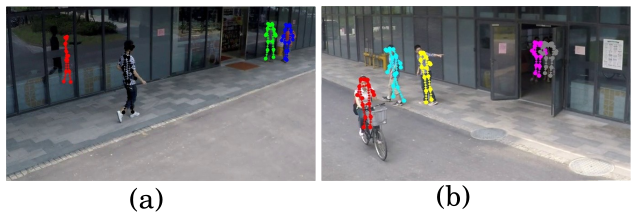}
\par\end{centering}
\caption{\label{fig:error_analysis}: Error mode examples. (a) Skeleton detection
inaccuracy: bad detection of the person reflected in the glass lead
to bad predictions by MPED-RNN. (b) Confusion in feature space: The
person riding the bicycle (red) has a moving pattern ``similar''
to a person walking.}
\end{figure}
Even though MPED-RNN outperforms related methods, it still makes incorrect
decisions. To understand the weaknesses of MPED-RNN, we sorted the
test sequences by decreasing level of error made by MPED-RNN and looked
for the root causes of the major mistakes. 

The most prominent source of errors is from the inaccuracy of the
skeleton detection and tracking. All of the skeleton detection methods
we tried produced inaccurate skeletons in several common difficult
cases such as low resolution of human area or unwanted lighting, contrast
or shadow. Moreover, when there is occlusion or multiple people crossing
each other, the tracking IDs can get lost or swapped and confuse MPED-RNN.
Figure \ref{fig:error_analysis}.a shows an example frame containing
a badly detected skeleton.

Apart from the input noise, a small portion of error comes from a
more interesting phenomenon when the abnormal action of subjects produce
similar skeletons to normal ones. Figure \ref{fig:error_analysis}.b
shows the case of a person slowly riding a bicycle with motion and
posture similar to walking, which tricks our model into a false negative.
This issue is a predicted downside of geometrical skeleton features,
where all appearance features have been filtered out. Augmenting the
skeleton structure with visual features is a future work towards solving
this issue.

\subsection{CUHK Avenue dataset}

We also tested MPED-RNN on the CUHK Avenue dataset, which is another
representative dataset for video anomaly detection. It contains 16
training videos and 21 testing videos captured from a single camera.
Based on earlier error analysis on the ShanghaiTech dataset, we understand
that the unstable skeleton inputs are the most important source of
inaccuracy. To avoid this issue, we manually leave out a set of video
frames where the main anomalous event is non-human related, or the
person involved is non-visible (\eg person out of view throwing an
object into the scene), or the main subject cannot be detected and
tracked. This selection is detailed in the Appendix A. We called the
remaining dataset HR-Avenue. On HR-Avenue, we achieved a frame-level
ROC AUC of 0.863, against 0.862 and 0.848 achieved by Liu \etal \cite{liu2018anopred}
and Conv-AE \cite{hasan2016learning}, respectively.

\section{Discussion}

With less than a hundred dimensions per frame on average, equal to
a small fraction of the popular visual features for anomaly detection
(ResNet features of 2048 \cite{hasan2016learning}, AlexNet fc7 of
4096 \cite{hinami2017joint}), skeleton features still provide equal
or better performance than current state-of-the-art methods. This
revives the hope for using semantic guided stage-by-stage approaches
for anomaly detection amid the trend of end-to-end image processing
deep networks. It also reflects the current trend of architectures
being modular, with multiple independent modules \cite{andreas2016neural,liu2018temporal}.
Apparently, MPED-RNN's performance still depends on the quality of
skeleton detection and tracking. This problem becomes more significant
in the case of low quality videos. It prevents us from trying our
method on UCSD Ped1/Ped2 \cite{mahadevan2010anomaly}, another popular
dataset whose video quality is too low to detect skeletons. Furthermore,
for the cases where skeletons are unavailable, appearance based features
can provide complementary information to help. This opens a promising
direction of combining these features in a cascaded model, where they
can cover the weaknesses of each other. Our message-passing scheme
can naturally be extended to incorporate sub-processes with non-skeleton
features.

Although dynamic movement and posture of single person can reflect
the anomalies in most cases, they do not contain information about
the interactions between multiple people in the events, and between
humans and objects. The global-local decomposition used in our method
can be extended to objects by exploring the part-based configuration
for each type of them. Towards multi-person/object anomalies, the
message passing framework in MPED-RNN is ready to extend support to
them, by expanding to inter-entity messages.

\section{Conclusions}

Through our experiments, we learned that skeleton motion sequences
are effective to identify human-related video anomalous events. We
observed that the decomposition of the skeleton sequence into global
movement and local deformation -- combined with our novel message-passing
encoder-decoder RNN architecture -- appropriately separates anomalous
sequences from normal sequences. MPED-RNN is simple, achieves competitive
performance and is highly interpretable. Future work includes examining
the regularity of inter-human interactions, combining skeleton features
with appearance counterparts, and expanding the component based model
to non-human objects.

{\small{}\bibliographystyle{ieee}
\bibliography{arxiv}

\begin{thebibliography}{10}\itemsep=-1pt

\bibitem{alahi2016social}
Alexandre Alahi, Kratarth Goel, Vignesh Ramanathan, Alexandre Robicquet, Li
  Fei-Fei, and Silvio Savarese.
\newblock Social {LSTM}: Human trajectory prediction in crowded spaces.
\newblock In {\em {IEEE} Conference on Computer Vision and Pattern
  Recognition}, pages 961--971, 2016.

\bibitem{andreas2016neural}
Jacob Andreas, Marcus Rohrbach, Trevor Darrell, and Dan Klein.
\newblock Neural module networks.
\newblock In {\em IEEE Conference on Computer Vision and Pattern Recognition},
  pages 39--48, 2016.

\bibitem{bau2017network}
David Bau, Bolei Zhou, Aditya Khosla, Aude Oliva, and Antonio Torralba.
\newblock Network dissection: Quantifying interpretability of deep visual
  representations.
\newblock In {\em {IEEE} Conference on Computer Vision and Pattern
  Recognition}, pages 3319--3327, 2017.

\bibitem{cho2014learning}
Kyunghyun Cho, Bart van Merrienboer, Caglar Gulcehre, Dzmitry Bahdanau, Fethi
  Bougares, Holger Schwenk, and Yoshua Bengio.
\newblock Learning phrase representations using {RNN} {Encoder--Decoder} for
  statistical machine translation.
\newblock In {\em Conference on Empirical Methods in Natural Language
  Processing}, pages 1724--1734, 2014.

\bibitem{chong2017abnormal}
Yong~Shean Chong and Yong~Haur Tay.
\newblock Abnormal event detection in videos using spatiotemporal autoencoder.
\newblock In {\em International Symposium on Neural Networks}, pages 189--196.
  Springer, 2017.

\bibitem{cocsar2017toward}
Serhan Co{\c{s}}ar, Giuseppe Donatiello, Vania Bogorny, Carolina Garate,
  Luis~Otavio Alvares, and Fran{\c{c}}ois Br{\'e}mond.
\newblock Toward abnormal trajectory and event detection in video surveillance.
\newblock {\em IEEE Transactions on Circuits and Systems for Video Technology},
  27(3):683--695, 2017.

\bibitem{du2016representation}
Yong Du, Yun Fu, and Liang Wang.
\newblock Representation learning of temporal dynamics for skeleton-based
  action recognition.
\newblock {\em IEEE Transactions on Image Processing}, 25(7):3010--3022, 2016.

\bibitem{elaoud2017analysis}
Amani Elaoud, Walid Barhoumi, Hassen Drira, and Ezzeddine Zagrouba.
\newblock Analysis of skeletal shape trajectories for person re-identification.
\newblock In {\em International Conference on Advanced Concepts for Intelligent
  Vision Systems}, pages 138--149. Springer, 2017.

\bibitem{fang2017rmpe}
Hao-Shu Fang, Shuqin Xie, Yu-Wing Tai, and Cewu Lu.
\newblock {RMPE}: Regional multi-person pose estimation.
\newblock In {\em {IEEE} International Conference on Computer Vision}, pages
  2353--2362, 2017.

\bibitem{fragkiadaki2015recurrent}
Katerina Fragkiadaki, Sergey Levine, Panna Felsen, and Jitendra Malik.
\newblock Recurrent network models for human dynamics.
\newblock In {\em IEEE International Conference on Computer Vision}, pages
  4346--4354, 2015.

\bibitem{greff2017lstm}
Klaus Greff, Rupesh~K Srivastava, Jan Koutnik, Bas~R Steunebrink, and Jurgen
  Schmidhuber.
\newblock {LSTM}: A search space odyssey.
\newblock {\em IEEE transactions on neural networks and learning systems},
  28(10):2222--2232, 2017.

\bibitem{gupta2018social}
Agrim Gupta, Justin Johnson, Li Fei-Fei, Silvio Savarese, and Alexandre Alahi.
\newblock Social gan: Socially acceptable trajectories with generative
  adversarial networks.
\newblock In {\em IEEE Conference on Computer Vision and Pattern Recognition},
  2018.

\bibitem{hasan2016learning}
Mahmudul Hasan, Jonghyun Choi, Jan Neumann, Amit~K Roy-Chowdhury, and Larry~S
  Davis.
\newblock Learning temporal regularity in video sequences.
\newblock In {\em IEEE Conference on Computer Vision and Pattern Recognition},
  pages 733--742, 2016.

\bibitem{hinami2017joint}
Ryota Hinami, Tao Mei, and Shin'ichi Satoh.
\newblock Joint detection and recounting of abnormal events by learning deep
  generic knowledge.
\newblock In {\em {IEEE} International Conference on Computer Vision}, pages
  3639--3647, 2017.

\bibitem{kim2009observe}
Jaechul Kim and Kristen Grauman.
\newblock Observe locally, infer globally: a space-time mrf for detecting
  abnormal activities with incremental updates.
\newblock In {\em IEEE Conference on Computer Vision and Pattern Recognition},
  pages 2921--2928, 2009.

\bibitem{kuhn1955hungarian}
Harold~W Kuhn.
\newblock The hungarian method for the assignment problem.
\newblock {\em Naval research logistics quarterly}, 2(1-2):83--97, 1955.

\bibitem{liu2018temporal}
Bingbin Liu, Serena Yeung, Edward Chou, De-An Huang, Li Fei-Fei, and
  Juan~Carlos Niebles.
\newblock Temporal modular networks for retrieving complex compositional
  activities in videos.
\newblock In {\em European Conference on Computer Vision}, pages 569--586.
  Springer, 2018.

\bibitem{liu2018anopred}
Wen Liu, Weixin Luo, Dongze Lian, and Shenghua Gao.
\newblock Future frame prediction for anomaly detection -- a new baseline.
\newblock In {\em IEEE Conference on Computer Vision and Pattern Recognition},
  2018.

\bibitem{lu2013abnormal}
Cewu Lu, Jianping Shi, and Jiaya Jia.
\newblock Abnormal event detection at 150 {FPS} in {MATLAB}.
\newblock In {\em {IEEE} International Conference on Computer Vision}, pages
  2720--2727, 2013.

\bibitem{luo2017revisit}
Weixin Luo, Wen Liu, and Shenghua Gao.
\newblock A revisit of sparse coding based anomaly detection in stacked {RNN}
  framework.
\newblock In {\em {IEEE} International Conference on Computer Vision}, pages
  341--349, 2017.

\bibitem{mahadevan2010anomaly}
Vijay Mahadevan, Weixin Li, Viral Bhalodia, and Nuno Vasconcelos.
\newblock Anomaly detection in crowded scenes.
\newblock In {\em IEEE Conference on Computer Vision and Pattern Recognition},
  pages 1975--1981, 2010.

\bibitem{srivastava2015unsupervised}
Nitish Srivastava, Elman Mansimov, and Ruslan Salakhudinov.
\newblock Unsupervised learning of video representations using {LSTMs}.
\newblock In {\em International Conference on Machine Learning}, pages
  843--852, 2015.

\bibitem{villegas2017learning}
Ruben Villegas, Jimei Yang, Yuliang Zou, Sungryull Sohn, Xunyu Lin, and Honglak
  Lee.
\newblock Learning to generate long-term future via hierarchical prediction.
\newblock In {\em International Conference on Machine Learning}, pages
  3560--3569, 2017.

\bibitem{vu2017energy}
Hung Vu, Tu~Dinh Nguyen, Anthony Travers, Svetha Venkatesh, and Dinh Phung.
\newblock Energy-based localized anomaly detection in video surveillance.
\newblock In {\em Pacific-Asia Conference on Knowledge Discovery and Data
  Mining}, pages 641--653. Springer, 2017.

\bibitem{xu2017detecting}
Dan Xu, Yan Yan, Elisa Ricci, and Nicu Sebe.
\newblock Detecting anomalous events in videos by learning deep representations
  of appearance and motion.
\newblock {\em Computer Vision and Image Understanding}, 156:117--127, 2017.

\bibitem{yuan2015online}
Yuan Yuan, Jianwu Fang, and Qi Wang.
\newblock Online anomaly detection in crowd scenes via structure analysis.
\newblock {\em IEEE Transactions on Cybernetics}, 45(3):548--561, 2015.

\bibitem{zhao2017spatio}
Yiru Zhao, Bing Deng, Chen Shen, Yao Liu, Hongtao Lu, and Xian-Sheng Hua.
\newblock Spatio-temporal autoencoder for video anomaly detection.
\newblock In {\em Proceedings of the 2017 ACM on Multimedia Conference}, pages
  1933--1941. ACM, 2017.

\bibitem{zimek2012survey}
Arthur Zimek, Erich Schubert, and Hans-Peter Kriegel.
\newblock A survey on unsupervised outlier detection in high-dimensional
  numerical data.
\newblock {\em Statistical Analysis and Data Mining: The ASA Data Science
  Journal}, 5(5):363--387, 2012.

\end{thebibliography}
}{\small\par}

\appendix

\section{HR-ShanghaiTech and HR-Avenue Datasets}

The ShanghaiTech dataset contains anomalies that are not related to
humans in 6 out of its 107 test videos. These 6 videos are:
\begin{itemize}
\item Camera 01: Videos 0130, 0135 and 0136;
\item Camera 06: Videos 0144 and 0145;
\item Camera 12: Video 0152.
\end{itemize}
HR-ShanghaiTech was assembled by removing those videos from the ShanghaiTech
dataset.

As for the HR-Avenue dataset, since the original Avenue dataset contains
only 21 testing videos, we ignored segments of the videos where the
anomalies were not detectable by the pose detector we employed or
where the anomaly was not related to a human. The ignored segments
were:
\begin{itemize}
\item Video 01: Frames 75 to 120, 390 to 436, 864 to 910 and 931 to 1000;
\item Video 02: Frames 272 to 319 and 723 to 763;
\item Video 03: Frames 293 to 340;
\item Video 06: Frames 561 to 624 and 814 to 1006;
\item Video 16: Frames 728 to 739.
\end{itemize}

\section{Code}

The source code for the models described in this paper is available
in the following link: https://github.com/RomeroBarata/skeleton\_based\_anomaly\_detection.

\end{document}